\documentclass{article}

\usepackage[numbers]{natbib}
\usepackage[preprint]{neurips_2022}

\usepackage[dvipsnames]{xcolor}         
\definecolor{linkColor}{rgb}{0.18,0.39,0.62}
\usepackage[utf8]{inputenc} 
\usepackage[T1]{fontenc}    
\usepackage[colorlinks=true,linkcolor=linkColor,citecolor=linkColor,filecolor=linkColor,urlcolor=linkColor]{hyperref}       
\usepackage{url}            
\usepackage{booktabs}       
\usepackage{amsfonts}       
\usepackage{nicefrac}       
\usepackage{microtype}      

\usepackage{graphicx}
\usepackage{arydshln}
\usepackage{booktabs}
\usepackage{multirow}
\usepackage{caption}
\usepackage{subcaption}
\usepackage{makecell}

\RequirePackage{algorithm}
\RequirePackage{algorithmic}

\usepackage{multirow}
\usepackage{amsmath}
\usepackage{capt-of}
\usepackage{tabularx}
\usepackage{epsfig}
\usepackage{amssymb}
\usepackage{amsfonts}
\usepackage{booktabs}
\usepackage{scalerel}
\usepackage[inline]{enumitem}
\usepackage{listings}
\usepackage{varwidth}
\usepackage[export]{adjustbox}
\usepackage{tikz}
\usetikzlibrary{tikzmark}

\usepackage{stmaryrd}
\usepackage{bbm}
\usepackage{wrapfig}
\usepackage{pifont}

\definecolor{deepblue}{rgb}{0,0,0.5}
\definecolor{officeblue}{RGB}{0,102,204}
\definecolor{deepred}{rgb}{0.6,0,0}
\definecolor{deepgreen}{rgb}{0,0.5,0}
\definecolor{mybrickred}{RGB}{182,50,28}

\definecolor{fillcolor}{RGB}{216,217,252}

\usepackage{amsmath,amsfonts,bm}

\def\eqref#1{equation~\ref{#1}}

\def\1{\bm{1}}

\def\vk{{\bm{k}}}

\def\vq{{\bm{q}}}

\DeclareMathAlphabet{\mathsfit}{\encodingdefault}{\sfdefault}{m}{sl}
\SetMathAlphabet{\mathsfit}{bold}{\encodingdefault}{\sfdefault}{bx}{n}

\DeclareMathOperator*{\argmax}{arg\,max}

\usepackage{pifont}
\title{Structured Prompting: \\ Scaling In-Context Learning to 1,000 Examples}

\author{\\
\textbf{Yaru Hao\thanks{~Equal contribution.},~\ Yutao Sun\footnotemark[1],~\ Li Dong,~Zhixiong Han,~Yuxian Gu,~Furu Wei} \\
Microsoft Research \\
\url{https://github.com/microsoft/LMOps} \\}

\date{}

\begin{document}
\maketitle
\begin{abstract}
Large language models have exhibited intriguing in-context learning capability, achieving promising zero- and few-shot performance without updating the parameters. However, conventional in-context learning is usually restricted by length constraints, rendering it ineffective to absorb supervision from a large number of examples. In order to go beyond few shots, we introduce \textit{structured prompting} that breaks the length limit and scales in-context learning to thousands of examples. Specifically, demonstration examples are separately encoded with well-designed position embeddings, and then they are jointly attended by the test example using a rescaled attention mechanism. So we can scale the number of exemplars with linear complexity instead of quadratic complexity with respect to length. Experimental results on a diverse set of tasks show that our approach improves end-task performance and reduces evaluation variance over conventional in-context learning as the number of demonstration examples increases. Code has been released at \url{https://aka.ms/structured-prompting}.
\end{abstract}

\section{Introduction}
\label{sec:intro}

In-context learning~\citep{gpt3} prompts pretrained language models to perform downstream tasks without any parameter update.
Rather than fine-tuning the parameters for few-shot learning, we feed task-specific instructions and input-output demonstrations into large language models.
Then the evaluation input is conditioned on the given context to make predictions.
The paradigm is appealing since we can host language modeling inference as a general-purpose service for a wide range of tasks.

Most previous studies~\citep{gpt3,gopher,mtnlg,metalm} of in-context learning are conducted for few-shot learning. For example, PaLM~\citep{palm} typically conditions on five demonstration examples for most benchmarks. However, the restricted number of training instances potentially limits the usage of in-context learning in practice, especially when we have many examples. In comparison, fine-tuning is able to consume much more training examples for supervision despite the costly training. The above data utilization issue motivates us to empower in-context learning with more demonstration examples.
Directly scaling up the size is challenging.
For example, the language models with absolute position embeddings are pretrained with a predefined length. So the naive concatenation of many examples typically exceeds the maximum length.
Moreover, the conventional self-attention mechanism suffers from quadratic complexity in terms of computation and memory consumption, rendering scaling up infeasible.
In addition, more shots tend to reduce the performance variance~\citep{zhao2021calibrate,lu2022order} caused by different choices and permutations of demonstration examples.

\begin{figure}[ht]
\includegraphics[width=\textwidth]{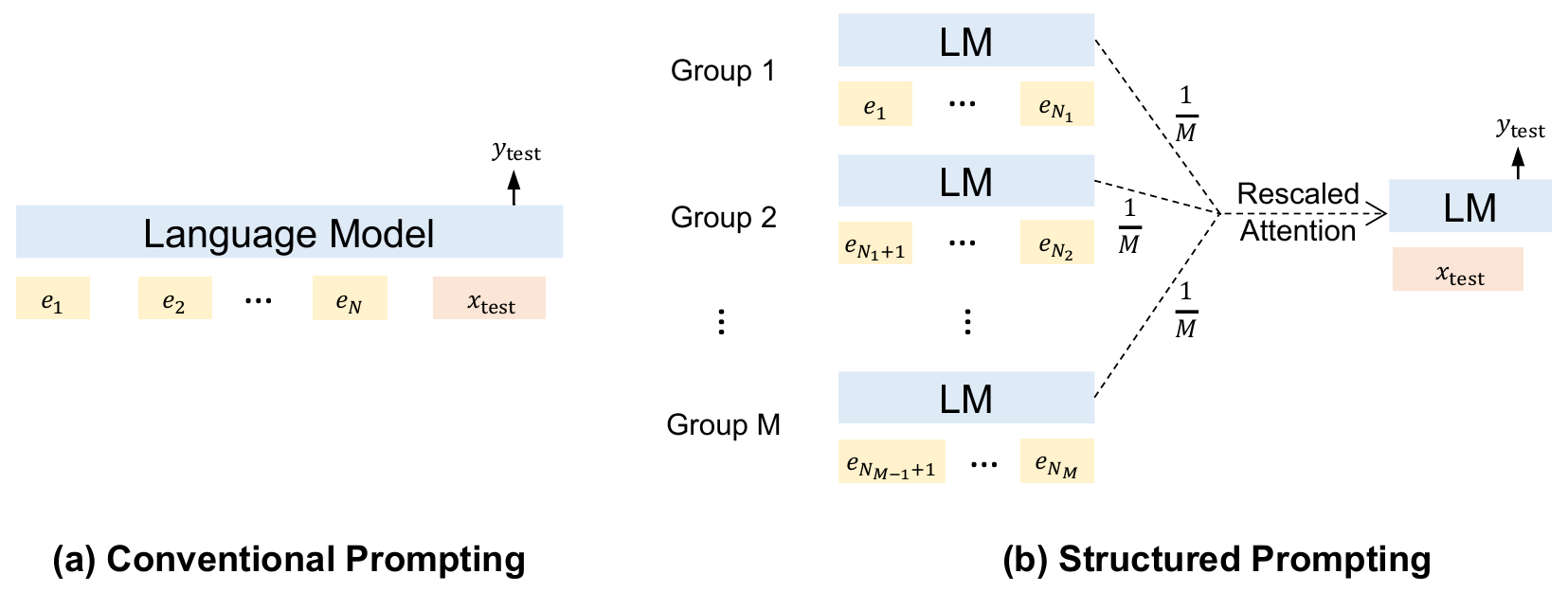}
\centering
\caption{The illustrations of conventional prompting and structured prompting. We first encode group-structured exemplars independently. Then all exemplars are incorporated into the test input through the rescaled attention mechanism.}
\label{fig:method}
\end{figure}

In this paper, we propose structured prompting to scale the number of examples to orders of magnitude larger and significantly improve stability.
Rather than simply concatenating all demonstrations together, we divide a large number of demonstrations into multiple groups, which are independently encoded by the language model.
So the encoding complexity becomes linear with respect to the number of groups, instead of quadratic complexity with respect to all examples.
The position embeddings of grouped prompts are right-aligned to be next to the test input.
Next, the input is encoded by conditioning on grouped prompts, where rescaled attention is proposed to normalize the attention scores.
Our structured prompting is flexible to encode plenty of context in an efficient way.
We conduct experiments on a variety of tasks, such as text classification, multi-choice, and open-ended tasks.
Structured prompting successfully scales the number of demonstrations to much larger sizes.
Our method substantially outperforms conventional in-context learning across various model sizes and tasks.
Moreover, the approach greatly improves the stability of in-context learning.

\section{Background: In-Context Learning}
\label{background}

In-context learning~\cite{gpt3} allows language models to recognize the desired task and generate answers for given inputs by conditioning on instructions and input-output demonstration examples, rather than updating model parameters as fine-tuning.

Formally, given a set of $N$ labeled examples $\mathcal{D}_\text{train}=\{(x_i,y_i)\}_{i=1}^N$ (i.e., $N$-shot in-context learning), each of them is transformed into a semantically meaningful demonstration $d_i = \mathcal{T}(x_i,y_i)$ using a hand-crafted template $\mathcal{T}$.
For example, the template of a binary sentiment classification task can be ``\emph{Sentence: $x_i$. Sentiment: $y_i$}'', where $y_i$ is \emph{Negative} or \emph{Positive}.
All demonstrations are concatenated as the context $\mathcal{Z}=d_1 \oplus ... \oplus d_N$.
Newlines or special tokens are used to delimit them.
For each test input $x_\text{test}$, we prompt the language model with the concatenation of $\mathcal{Z}$ and $x_\text{test}$.
The predicted answer is the completion with the highest language model probability, i.e., $\argmax_{c \in \mathcal{Y}} P_\text{LM}(y^c|\mathcal{Z} \oplus \mathcal{T}(x_\text{test}))$, where $\mathcal{Y}$ is the set of all possible candidates.
For conventional in-context learning, the number of demonstrations $N$ is restricted by the maximum length of pretrained Transformers (typically $2048$), which typically fits $5$ to $100$ examples depending on the datasets.

\section{Methods}
\label{method}

In this section, we introduce \textit{structured prompting}, which scales in-context learning to many examples under limited computation complexity.
An overview of our approach is shown in Figure~\ref{fig:method}.
First, we divide examples into groups. We obtain representations of group-structured exemplars independently with right-aligned position embeddings.
Second, we incorporate the encoding results into the test input through the rescaled attention mechanism in each layer. Then the language model generates the answer.

\subsection{Grouped Context Encoding}

Suppose we have $N$ demonstration examples.
We randomly divide these examples into $M$ groups $\{\mathcal{Z}_i\}_{i=1}^{M}$.
Each group is a concatenation of exemplars $\mathcal{Z}_i = d_{N_{i-1}+1} \oplus ... \oplus d_{N_i}$, where $N_0=0$ and $N_M=N$.
As shown in Figure~\ref{fig:method}, all exemplar groups are separately encoded by the language model.
Then we use the context encoding results for structured prompting. Notice that only key and value vectors of self-attention need to be cached, which are attended by the test input.

Grouped context encoding is able to consume longer sequences.
In contrast, conventional in-context learning cannot exploit context efficiently because the concatenation of all examples far exceeds the window size of pretrained Transformers.
Moreover, the computation complexity of conventional in-context learning is quadratic to the number of demonstration examples $N$ because of the matrix multiplication of queries and keys.
It is infeasible when $N$ increases.
Our approach improves it through splitting groups, which reduces the complexity from $\mathcal{O}(N^2)$ to $\mathcal{O}(N^2/M)$.

\paragraph{Right-Aligned Position Embedding}
We right-align all the groups so that they have the same maximum position index.
Hence all groups can have the same relative distance with respect to the test input.
It is critical that the test input can be adjacent to all exemplars and pay equal attention to them.
One way is to use left padding, i.e., pad tokens or space tokens.
The other way is to set a maximum length for grouped context, truncating exemplars from the left side.

\subsection{Structured Prompting}

After encoding context exemplars in groups, the next step is to use them for prompting.
As shown in Figure~\ref{fig:method}, all exemplars are incorporated into representations of the test input through a rescaled attention mechanism.
Specifically, the test input is fed into the language model, conditioning on both itself and grouped exemplars.
Let $L$ denote the maximum length of grouped context.
The position index of the test input starts with $L+1$ so that it is contiguous to all groups.

\paragraph{Rescaled Attention}
We use $x$ instead of $\mathcal{T}(x_\text{test})$ for brevity.
In each layer, we concatenate the keys and values of all exemplars and the test input, i.e., $\hat{K}=[K_{\mathcal{Z}_1}, ..., K_{\mathcal{Z}_M}, K_x], \hat{V}=[V_{\mathcal{Z}_1}, ..., V_{\mathcal{Z}_M}, V_x]$.
The test input $x$ attends both demonstrations and itself with causal masks.
Then the attention output is computed via:
\begin{align}
{\rm Attention}&(Q_{x}, \hat{K}, \hat{V}) = A \hat{V}
\\
A_{ij} &\propto
\begin{cases}
M {\rm exp}(\frac{\vq_i \cdot \vk_j}{\sqrt{d}} ) & j \in x \\
{\rm exp}(\frac{\vq_i \cdot \vk_j}{\sqrt{d}} ) & j \in \mathcal{Z}_1, \dots, \mathcal{Z}_M
\end{cases} \label{eq:rescaled:attention}
\end{align}
where $\sum_{j}{A_{ij}}=1$, the query vector $\vq_i \in Q_x$, the key vector $\vk_j \in \hat{K}^{\intercal}$, and $d$ is dimension of queries and keys.

Compared with vanilla self-attention used in Transformers~\cite{transformer}, the only difference is the scaling factor $M$ in Equation~(\ref{eq:rescaled:attention}).
Without rescaled attention, the test input will attend too much to exemplars and ignore itself as the number of exemplars increases.
Intuitively, our method modifies the softmax function in self-attention by repeating test input tokens $M$ times.
So we can augment the test input with multiple groups of context.

\section{Experiments}
\label{experiments}

\subsection{Setup}
\label{setup}

\paragraph{Models}
We conduct experiments on open-source GPT-like (i.e., decoder-only Transformer) models released by \cite{fairseqlm}.
We use three models of different sizes with 1.3B, 6.7B, and 13B parameters.
The context window contains up to 2048 tokens.
For large-scale experiments, we use BLOOM-176B~\citep{bloom}.

\paragraph{Datasets}
We evaluate structured prompting on a wide range of tasks grouped into text classification, multi-choice, and open-ended generation tasks.
For text classification, we use datasets of sentiment: SST-2~\citep{sst}, SST-5~\citep{sst}, MR~\citep{mr}, Subj~\citep{subj}; topic: DBPedia~\citep{dbpedia:agnews}, AGNews~\citep{dbpedia:agnews}, TREC~\citep{trec}; natural language inference: CB~\citep{cb}, RTE~\citep{rte1,rte2,rte3,rte5}; and question answering: BoolQ~\citep{boolq}.
For multi-choice tasks, we consider sentence completion: HellaSwag~\citep{hellaswag}, StoryCloze~\citep{storycloze}; commensense reasoning: PIQA~\citep{piqa}, OpenBookQA~\citep{obqa}, ARC-Easy~\citep{arc}, ARC-Challenge~\citep{arc}; and COPA from SuperGLUE benchmark~\citep{superglue}.
For open-ended generation, we consider closed-book question answering: NaturalQS~\citep{nq}, WebQS~\citep{webqs}, TriviaQA~\citep{triviaqa}; and extractive reading comprehension: SQuAD~\citep{squad1}, SQuADv2~\citep{squad2}.

\paragraph{Evaluation Protocol}
Following~\citep{gpt3}, we randomly draw $N$ fixed examples from the training set as conditioning and report evaluation results on the development set.
The demonstrations are separated by a special token.
For StoryCloze, there is no available training set so we draw from the development set and evaluate on the test set.
To reduce cost, we use 4k test examples for inference.
There are only six datasets with development sets larger than 4k and we randomly sample a fixed subset of them.

We design a hand-crafted template for each text classification dataset.
For other datasets, we follow the same template in GPT-3.
All templates are listed in Appendix~\ref{app:template}.
Notice that demonstrations for reading comprehension datasets (SQuAD, SQuADv2) are constructed slightly differently from the original GPT-3.
In GPT-3, the demonstrations provided for each test input are question-answer pairs from the same background passage as it.
Here we consider a more strict setting where demonstrations are constructed with different passage-question-answer combinations from the training set.

For multi-choice tasks, we score each completion by the per-token language model likelihood (normalize perplexity by length) and pick the one with the highest score as the final answer.
For text classification, we treat it as a multi-choice task with only one token per option and design meaningful names for each option.
For open-ended generation tasks, we use beam search with a beam width of 3, a length penalty of $\alpha=0.6$, and a maximum generation length of 30.
We report exact-match accuracy for closed-book QA and F1 score for SQuAD and SQuADv2.

For conventional prompting, we report results for 0-shot and the largest shot that fills one context window ($1\times$). 
The maximum number of shots is calculated based on the average length of each dataset.
Structured prompting is no longer limited by the context window size and can scale in-context learning to thousands of examples.
For datasets with shorter lengths (e.g., SST-2, Subj), we report results of 500-shot and 1000-shot. 
For datasets with longer lengths (e.g., AGNews, SQuAD), we choose the number of shots according to their average lengths.
We find it beneficial to put as many demonstrations as possible in each group.
Thus we adopt it in our main experiments.
Under each setting, we use six different random seeds for all tasks and report the mean and variance.

\begin{table}
\scalebox{0.92}{
\begin{tabular}{@{}l@{\hspace{0.8\tabcolsep}}c @{\hspace{0.8\tabcolsep}} c@{\hspace{0.8\tabcolsep}}c@{}}
\toprule
\textbf{SST-2} & \multicolumn{3}{c}{Model Size} \\
$N$ & 1.3B & 6.7B & 13B \\
\midrule
0 & 70.5$_{0.0}$ & 65.9$_{0.0}$ & 64.7$_{0.0}$ \\
102$(\!1\!\times\!)$ & 88.0$_{7.1}$ & 92.2$_{1.5}$ & 93.8$_{0.7}$ \\
\midrule
500 & 89.6$_{6.4}$ & 93.3$_{0.9}$ & 93.5$_{0.8}$  \\
1000 &\textbf{92.2$_{0.5}$}& \textbf{93.3$_{0.6}$} &\textbf{94.1$_{0.4}$} \\
\bottomrule
\\ \addlinespace[0.25ex]
\end{tabular}
}
\hspace{1.1em}
\scalebox{0.92}{
\begin{tabular}{@{}l@{\hspace{0.8\tabcolsep}}c @{\hspace{0.8\tabcolsep}} c@{\hspace{0.8\tabcolsep}}c@{}}
\toprule
\textbf{SST-5} & \multicolumn{3}{c}{Model Size} \\$N$ & 1.3B & 6.7B & 13B \\
\midrule
0 & 39.2$_{0.0}$ & 32.5$_{0.0}$ & 37.6$_{0.0}$ \\
66$(\!1\!\times\!)$ & 38.0$_{5.5}$ & 46.1$_{2.6}$ & 47.2$_{2.6}$ \\
\midrule
500 & 42.6$_{1.1}$ & 49.2$_{0.7}$ &\textbf{50.1$_{1.2}$} \\
1000 & \textbf{42.6$_{0.5}$} &\textbf{49.3$_{0.4}$}& 48.7$_{0.6}$  \\
\bottomrule
\\ \addlinespace[0.25ex]
\end{tabular}
}
\hspace{1.1em}
\scalebox{0.92}{
\begin{tabular}{@{}l@{\hspace{0.8\tabcolsep}}c @{\hspace{0.8\tabcolsep}} c@{\hspace{0.8\tabcolsep}}c@{}}
\toprule
\textbf{MR} & \multicolumn{3}{c}{Model Size} \\$N$ & 1.3B & 6.7B & 13B \\
\midrule
0 & 65.9$_{0.0}$ & 58.0$_{0.0}$ & 60.7$_{0.0}$ \\
63$(\!1\!\times\!)$ & 84.1$_{4.1}$ & 91.8$_{0.7}$ & 92.7$_{0.4}$ \\
\midrule
500 & 88.5$_{2.1}$ &\textbf{92.2$_{0.2}$}&\textbf{92.9$_{0.2}$} \\
1000 &\textbf{89.7$_{0.3}$}& \textbf{92.2$_{0.2}$} & 92.6$_{0.3}$  \\
\bottomrule
\\ \addlinespace[0.25ex]
\end{tabular}
}

\scalebox{0.92}{
\begin{tabular}{@{}l@{\hspace{0.8\tabcolsep}}c @{\hspace{0.8\tabcolsep}} c@{\hspace{0.8\tabcolsep}}c@{}}
\toprule
\textbf{Subj} & \multicolumn{3}{c}{Model Size} \\$N$ & 1.3B & 6.7B & 13B \\
\midrule
0 & 72.5$_{0.0}$ & 63.4$_{0.0}$ & 76.2$_{0.0}$ \\
55$(\!1\!\times\!)$ & 85.3$_{6.8}$ & 67.1$_{7.0}$ & 88.4$_{3.2}$ \\
\midrule
500 & 88.9$_{0.7}$ & 65.2$_{2.7}$ &\textbf{90.2$_{1.2}$} \\
1000 &\textbf{89.5$_{0.8}$}&\textbf{69.3$_{1.5}$}& 89.3$_{1.2}$   \\
\bottomrule
\\ \addlinespace[0.25ex]
\end{tabular}
}
\hspace{0.05em}
\scalebox{0.92}{
\begin{tabular}{@{}l@{\hspace{0.8\tabcolsep}}c @{\hspace{0.8\tabcolsep}} c@{\hspace{0.8\tabcolsep}}c@{}}
\toprule
\textbf{DBPedia} & \multicolumn{3}{c}{Model Size} \\$N$ & 1.3B & 6.7B & 13B \\
\midrule
0 & 76.6$_{0.0}$ & 77.3$_{0.0}$ & 79.8$_{0.0}$ \\
21$(\!1\!\times\!)$ & 84.7$_{5.6}$ & 68.2$_{3.9}$ & 91.5$_{3.1}$ \\
\midrule
100 &\textbf{87.8$_{2.0}$}& 93.8$_{0.6}$ & 94.2$_{0.8}$  \\
200 & 87.2$_{2.1}$ &\textbf{94.3$_{0.4}$}&\textbf{94.9$_{0.2}$} \\
\bottomrule
\\ \addlinespace[0.25ex]
\end{tabular}
}
\hspace{0.05em}
\scalebox{0.92}{
\begin{tabular}{@{}l@{\hspace{0.8\tabcolsep}}c @{\hspace{0.8\tabcolsep}} c@{\hspace{0.8\tabcolsep}}c@{}}
\toprule
\textbf{AGNews} & \multicolumn{3}{c}{Model Size} \\$N$ & 1.3B & 6.7B & 13B \\
\midrule
0 & 46.4$_{0.0}$ & 29.0$_{0.0}$ & 37.3$_{0.0}$ \\
23$(\!1\!\times\!)$ & 70.6$_{7.9}$ &\textbf{76.7$_{5.9}$}& 83.4$_{6.2}$  \\
\midrule
100 & 71.6$_{14.5}$ & 74.3$_{5.2}$ & 85.4$_{1.9}$ \\
200 &\textbf{76.2$_{9.0}$}& 75.7$_{3.2}$ &\textbf{86.3$_{1.3}$} \\
\bottomrule
\\ \addlinespace[0.25ex]
\end{tabular}
}

\scalebox{0.92}{
\begin{tabular}{@{}l@{\hspace{0.8\tabcolsep}}c @{\hspace{0.8\tabcolsep}} c@{\hspace{0.8\tabcolsep}}c@{}}
\toprule
\textbf{TREC} & \multicolumn{3}{c}{Model Size} \\$N$ & 1.3B & 6.7B & 13B \\
\midrule
0 & 46.4$_{0.0}$ & 41.6$_{0.0}$ & 44.6$_{0.0}$ \\
116$(\!1\!\times\!)$ & 62.3$_{3.5}$ & 75.8$_{2.0}$ & 81.4$_{3.5}$ \\
\midrule
500 & 66.5$_{1.2}$ & 79.3$_{1.0}$ & 84.4$_{2.2}$ \\
1000 &\textbf{66.6$_{1.9}$}&\textbf{79.4$_{0.9}$}&\textbf{85.1$_{0.8}$} \\
\bottomrule
\end{tabular}
}
\hspace{0.85em}
\scalebox{0.92}{
\begin{tabular}{@{}l@{\hspace{0.8\tabcolsep}}c @{\hspace{0.8\tabcolsep}} c@{\hspace{0.8\tabcolsep}}c@{}}
\toprule
\textbf{CB} & \multicolumn{3}{c}{Model Size} \\$N$ & 1.3B & 6.7B & 13B \\
\midrule
0 & 37.5$_{0.0}$ &\textbf{58.9$_{0.0}$}& 28.6$_{0.0}$  \\
18$(\!1\!\times\!)$ & 48.2$_{5.1}$ & 50.3$_{0.7}$ & 48.2$_{12.9}$ \\
\midrule
125 & 56.5$_{1.9}$ & 50.0$_{0.0}$ &\textbf{62.2$_{1.9}$} \\
250 &\textbf{56.9$_{2.8}$}& 50.0$_{0.0}$ & 59.8$_{0.9}$  \\
\bottomrule
\end{tabular}
}
\hspace{0.85em}
\scalebox{0.92}{
\begin{tabular}{@{}l@{\hspace{0.8\tabcolsep}}c @{\hspace{0.8\tabcolsep}} c@{\hspace{0.8\tabcolsep}}c@{}}
\toprule
\textbf{BoolQ} & \multicolumn{3}{c}{Model Size} \\$N$ & 1.3B & 6.7B & 13B \\
\midrule
0 & 59.5$_{0.0}$ &\textbf{64.2$_{0.0}$}& 67.0$_{0.0}$  \\
6$(\!1\!\times\!)$ & 57.4$_{8.0}$ & 63.8$_{6.6}$ & 70.2$_{2.2}$ \\
\midrule
50 & 59.3$_{7.1}$ & 61.7$_{8.8}$ & 72.5$_{2.6}$ \\
100 &\textbf{60.4$_{4.1}$}& 63.9$_{5.0}$ &\textbf{73.1$_{1.2}$} \\
\bottomrule
\end{tabular}
}
\vspace{0.2cm}
\caption{Results on text classification. $(\!1\!\times\!)$ is the maximum shot of conventional in-context learning.}
\label{tbl:classification}
\end{table}

\begin{table}
\scalebox{0.88}{
\begin{tabular}{@{}l@{\hspace{0.8\tabcolsep}}c @{\hspace{0.8\tabcolsep}} c@{\hspace{0.8\tabcolsep}}c@{}}
\toprule
\textbf{HellaSwag} & \multicolumn{3}{c}{Model Size} \\$N$ & 1.3B & 6.7B & 13B \\
\midrule
0 & 56.1$_{0.0}$ & 67.6$_{0.0}$ & 70.9$_{0.0}$ \\
24$(\!1\!\times\!)$ & 56.0$_{0.4}$ & 68.3$_{0.2}$ & 71.9$_{0.3}$ \\
\midrule
100 &\textbf{56.7$_{0.2}$}& 68.7$_{0.3}$ & 72.0$_{0.2}$  \\
200 & 56.4$_{0.3}$ &\textbf{68.8$_{0.2}$}&\textbf{72.1$_{0.2}$} \\
\bottomrule
\\ \addlinespace[0.25ex]
\end{tabular}
}
\hspace{0.01em}
\scalebox{0.88}{
\begin{tabular}{@{}l@{\hspace{0.8\tabcolsep}}c @{\hspace{0.8\tabcolsep}} c@{\hspace{0.8\tabcolsep}}c@{}}
\toprule
\textbf{StoryCloze} & \multicolumn{3}{c}{Model Size} \\$N$ & 1.3B & 6.7B & 13B \\
\midrule
0 & 77.0$_{0.0}$ & 79.3$_{0.0}$ & 80.2$_{0.0}$ \\
38$(\!1\!\times\!)$ & 78.6$_{0.8}$ & 83.2$_{0.5}$ & 85.0$_{0.4}$ \\
\midrule
500 &\textbf{78.7$_{0.2}$}&\textbf{83.6$_{0.1}$}&\textbf{86.1$_{0.2}$} \\
1000 & 78.3$_{0.2}$ & 83.5$_{0.1}$ & 85.7$_{0.2}$ \\
\bottomrule
\\ \addlinespace[0.25ex]
\end{tabular}
}
\hspace{0.01em}
\scalebox{0.88}{
\begin{tabular}{@{}l@{\hspace{0.8\tabcolsep}}c @{\hspace{0.8\tabcolsep}} c@{\hspace{0.8\tabcolsep}}c@{}}
\toprule
\textbf{ARC-E} & \multicolumn{3}{c}{Model Size} \\$N$ & 1.3B & 6.7B & 13B \\
\midrule
0 & 48.3$_{0.0}$ & 57.5$_{0.0}$ & 58.2$_{0.0}$ \\
70$(\!1\!\times\!)$ & 61.9$_{0.5}$ & 70.4$_{0.5}$ & 73.9$_{0.6}$ \\
\midrule
500 &\textbf{62.3$_{0.5}$}&\textbf{71.2$_{0.5}$}&\textbf{75.4$_{0.9}$} \\
1000 & 61.9$_{0.5}$ & 71.1$_{0.5}$ & 75.3$_{0.3}$ \\
\bottomrule
\\ \addlinespace[0.25ex]
\end{tabular}
}

\scalebox{0.72}{
\begin{tabular}{@{}l@{\hspace{0.7\tabcolsep}}c @{\hspace{0.7\tabcolsep}} c@{\hspace{0.7\tabcolsep}}c@{}}
\toprule
\textbf{PIQA} & \multicolumn{3}{c}{Model Size} \\$N$ & 1.3B & 6.7B & 13B \\
\midrule
0 & 74.3$_{0.0}$ & 76.8$_{0.0}$ & 78.1$_{0.0}$ \\
54$(\!1\!\times\!)$ &\textbf{74.4$_{0.5}$}& 78.0$_{0.5}$ & 79.8$_{0.4}$  \\
\midrule
500 & 74.3$_{0.2}$ &\textbf{78.6$_{0.1}$}& 79.9$_{0.1}$  \\
1000 & 74.1$_{0.1}$ & 78.3$_{0.3}$ &\textbf{80.3$_{0.2}$} \\
\bottomrule
\end{tabular}
}
\hspace{0.0001em}
\scalebox{0.72}{
\begin{tabular}{@{}l@{\hspace{0.7\tabcolsep}}c @{\hspace{0.7\tabcolsep}} c@{\hspace{0.7\tabcolsep}}c@{}}
\toprule
\textbf{OBQA} & \multicolumn{3}{c}{Model Size} \\$N$ & 1.3B & 6.7B & 13B \\
\midrule
0 & 34.4$_{0.0}$ & 35.8$_{0.0}$ & 39.0$_{0.0}$ \\
118$(\!1\!\times\!)$ & 34.2$_{0.8}$ & 39.6$_{0.4}$ & 41.1$_{1.1}$ \\
\midrule
500 &\textbf{35.0$_{0.7}$}&\textbf{41.0$_{1.0}$}& 42.7$_{0.7}$   \\
1000 & 34.5$_{0.5}$ & 40.9$_{0.7}$ &\textbf{43.0$_{0.6}$} \\
\bottomrule
\end{tabular}
}
\hspace{0.0001em}
\scalebox{0.72}{
\begin{tabular}{@{}l@{\hspace{0.7\tabcolsep}}c @{\hspace{0.7\tabcolsep}} c@{\hspace{0.7\tabcolsep}}c@{}}
\toprule
\textbf{ARC-C} & \multicolumn{3}{c}{Model Size} \\$N$ & 1.3B & 6.7B & 13B \\
\midrule
0 & 30.2$_{0.0}$ & 32.2$_{0.0}$ & 35.9$_{0.0}$ \\
61$(\!1\!\times\!)$ &\textbf{33.9$_{1.5}$}& 39.1$_{0.4}$ & 42.3$_{0.7}$  \\
\midrule
500 & 32.6$_{0.6}$ & 39.2$_{0.4}$ &\textbf{42.6$_{0.8}$} \\
1000 & 32.9$_{0.4}$ &\textbf{39.4$_{0.6}$}& 42.2$_{0.4}$  \\
\bottomrule
\end{tabular}
}
\hspace{0.0001em}
\scalebox{0.72}{
\begin{tabular}{@{}l@{\hspace{0.7\tabcolsep}}c @{\hspace{0.7\tabcolsep}} c@{\hspace{0.7\tabcolsep}}c@{}}
\toprule
\textbf{COPA} & \multicolumn{3}{c}{Model Size} \\$N$ & 1.3B & 6.7B & 13B \\
\midrule
0 & 71.0$_{0.0}$ & 83.0$_{0.0}$ & 81.0$_{0.0}$ \\
134$(\!1\!\times\!)$ & 75.2$_{1.6}$ &\textbf{85.7$_{1.4}$}& 87.3$_{2.6}$  \\
\midrule
200 & 75.5$_{1.7}$ & 85.7$_{1.7}$ & 86.8$_{2.4}$ \\
400 &\textbf{76.5$_{2.3}$}& 85.3$_{0.9}$ &\textbf{87.7$_{1.5}$} \\
\bottomrule
\end{tabular}
}
\vspace{0.2cm}
\caption{Results on multi-choice tasks. $(\!1\!\times\!)$ is the maximum shot of conventional in-context learning.}
\label{tbl:multichoice}
\end{table}

\begin{table}
\scalebox{0.92}{
\begin{tabular}{@{}l@{\hspace{0.8\tabcolsep}}c @{\hspace{0.8\tabcolsep}} c@{\hspace{0.8\tabcolsep}}c@{}}
\toprule
\textbf{NQ} & \multicolumn{3}{c}{Model Size} \\$N$ & 1.3B & 6.7B & 13B \\
\midrule
0 & 1.2$_{0.0}$ & 0.4$_{0.0}$ & 1.1$_{0.0}$ \\
100$(\!1\!\times\!)$ &\textbf{7.2$_{0.2}$}&\textbf{14.0$_{0.2}$}& 16.1$_{0.3}$   \\
\midrule
500 &\textbf{7.2$_{0.2}$}& 13.9$_{0.1}$ &\textbf{16.3$_{0.3}$} \\
1000 & 7.0$_{0.1}$ & 13.7$_{0.2}$ & 16.3$_{0.4}$ \\
\bottomrule
\\ \addlinespace[0.25ex]
\end{tabular}
}
\hspace{0.4em}
\scalebox{0.92}{
\begin{tabular}{@{}l@{\hspace{0.8\tabcolsep}}c @{\hspace{0.8\tabcolsep}} c@{\hspace{0.8\tabcolsep}}c@{}}
\toprule
\textbf{WebQS} & \multicolumn{3}{c}{Model Size} \\$N$ & 1.3B & 6.7B & 13B \\
\midrule
0 & 0.8$_{0.0}$ & 0.1$_{0.0}$ & 0.8$_{0.0}$ \\
108$(\!1\!\times\!)$ & 14.9$_{1.1}$ & 22.2$_{1.4}$ & 24.4$_{1.2}$ \\
\midrule
500 & 16.6$_{0.3}$ & 23.0$_{0.5}$ & 26.4$_{0.5}$ \\
1000 &\textbf{17.1$_{0.3}$}&\textbf{23.7$_{0.7}$}&\textbf{27.1$_{0.5}$} \\
\bottomrule
\\ \addlinespace[0.25ex]
\end{tabular}
}
\hspace{0.4em}
\scalebox{0.92}{
\begin{tabular}{@{}l@{\hspace{0.8\tabcolsep}}c @{\hspace{0.8\tabcolsep}} c@{\hspace{0.8\tabcolsep}}c@{}}
\toprule
\textbf{TriviaQA} & \multicolumn{3}{c}{Model Size} \\$N$ & 1.3B & 6.7B & 13B \\
\midrule
0 & 8.3$_{0.0}$ & 1.4$_{0.0}$ & 6.8$_{0.0}$ \\
66$(\!1\!\times\!)$ & 17.9$_{1.2}$ & 31.8$_{2.2}$ & 37.5$_{3.4}$ \\
\midrule
125 & 18.5$_{1.0}$ & 31.5$_{1.7}$ & 37.2$_{2.6}$ \\
250 &\textbf{19.5$_{0.6}$}&\textbf{33.0$_{1.1}$}&\textbf{39.9$_{1.4}$} \\
\bottomrule
\\ \addlinespace[0.25ex]
\end{tabular}
}

\scalebox{0.92}{
\begin{tabular}{@{}l@{\hspace{0.8\tabcolsep}}c @{\hspace{0.8\tabcolsep}} c@{\hspace{0.8\tabcolsep}}c@{}}
\toprule
\textbf{SQuAD} & \multicolumn{3}{c}{Model Size} \\$N$ & 1.3B & 6.7B & 13B \\
\midrule
0 & 24.1$_{0.0}$ & 25.6$_{0.0}$ & 31.6$_{0.0}$ \\
6$(\!1\!\times\!)$ & 52.4$_{3.0}$ & 67.0$_{1.2}$ & 68.7$_{1.5}$ \\
\midrule
25 & 53.8$_{2.5}$ & 68.2$_{0.8}$ & 71.8$_{0.9}$ \\
50 &\textbf{55.2$_{1.3}$}&\textbf{69.3$_{0.5}$}&\textbf{73.6$_{0.6}$} \\
\bottomrule
\end{tabular}
}
\hspace{0.001em}
\scalebox{0.92}{
\begin{tabular}{@{}l@{\hspace{0.8\tabcolsep}}c @{\hspace{0.8\tabcolsep}} c@{\hspace{0.8\tabcolsep}}c@{}}
\toprule
\textbf{SQuADv2} & \multicolumn{3}{c}{Model Size} \\$N$ & 1.3B & 6.7B & 13B \\
\midrule
0 & 11.3$_{0.0}$ & 12.0$_{0.0}$ & 14.8$_{0.0}$ \\
6$(\!1\!\times\!)$ & 30.3$_{3.0}$ & 36.8$_{3.4}$ & 36.2$_{2.2}$ \\
\midrule
25 & 31.6$_{2.1}$ & 37.9$_{2.4}$ & 37.6$_{1.2}$ \\
50 &\textbf{32.0$_{1.4}$}&\textbf{38.3$_{1.0}$}&\textbf{38.3$_{0.9}$} \\
\bottomrule
\end{tabular}
}
\vspace{0.2cm}
\caption{Results on open-ended tasks. $(\!1\!\times\!)$ is the maximum shot of conventional in-context learning.}
\label{tbl:generation}
\end{table}

\begin{figure}[h!]
\begin{subfigure}{\textwidth}
    \centering
    \includegraphics[width=\textwidth]{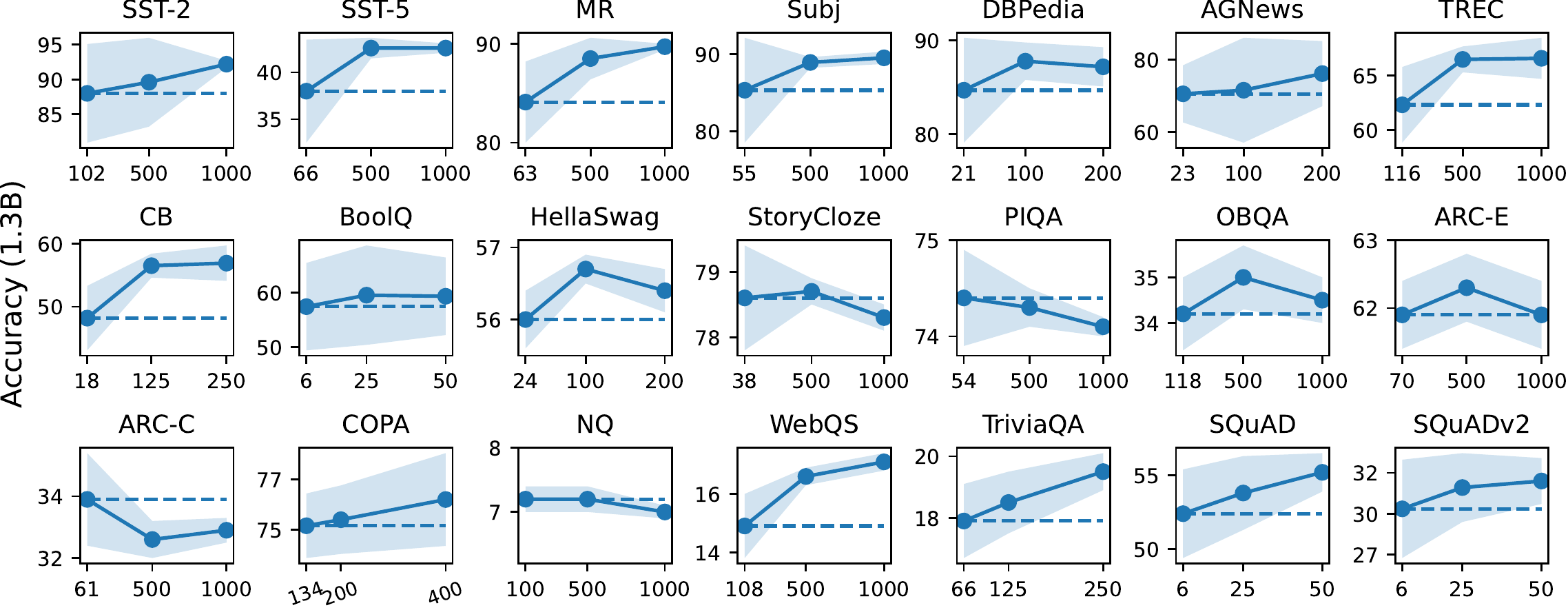}
\end{subfigure}
\begin{subfigure}{\textwidth}
    \centering
    \includegraphics[width=\textwidth]{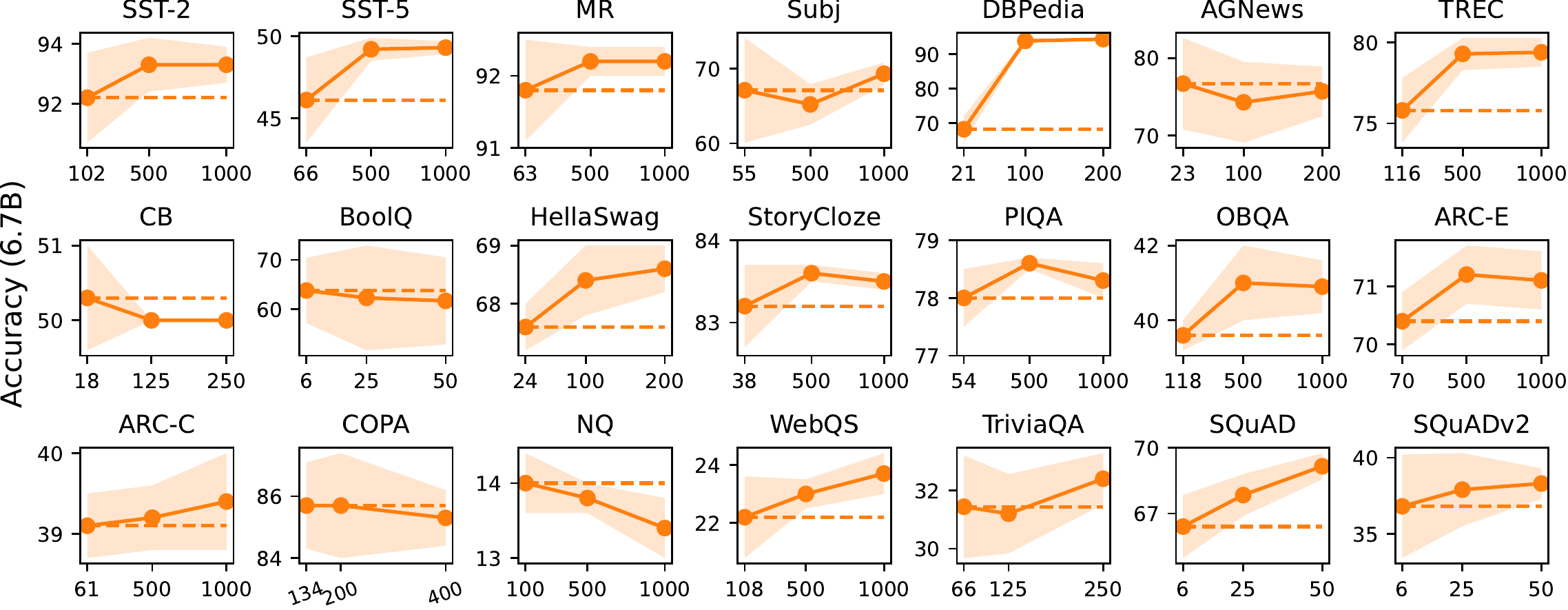}
\end{subfigure}
\begin{subfigure}{\textwidth}
    \centering
    \includegraphics[width=\textwidth]{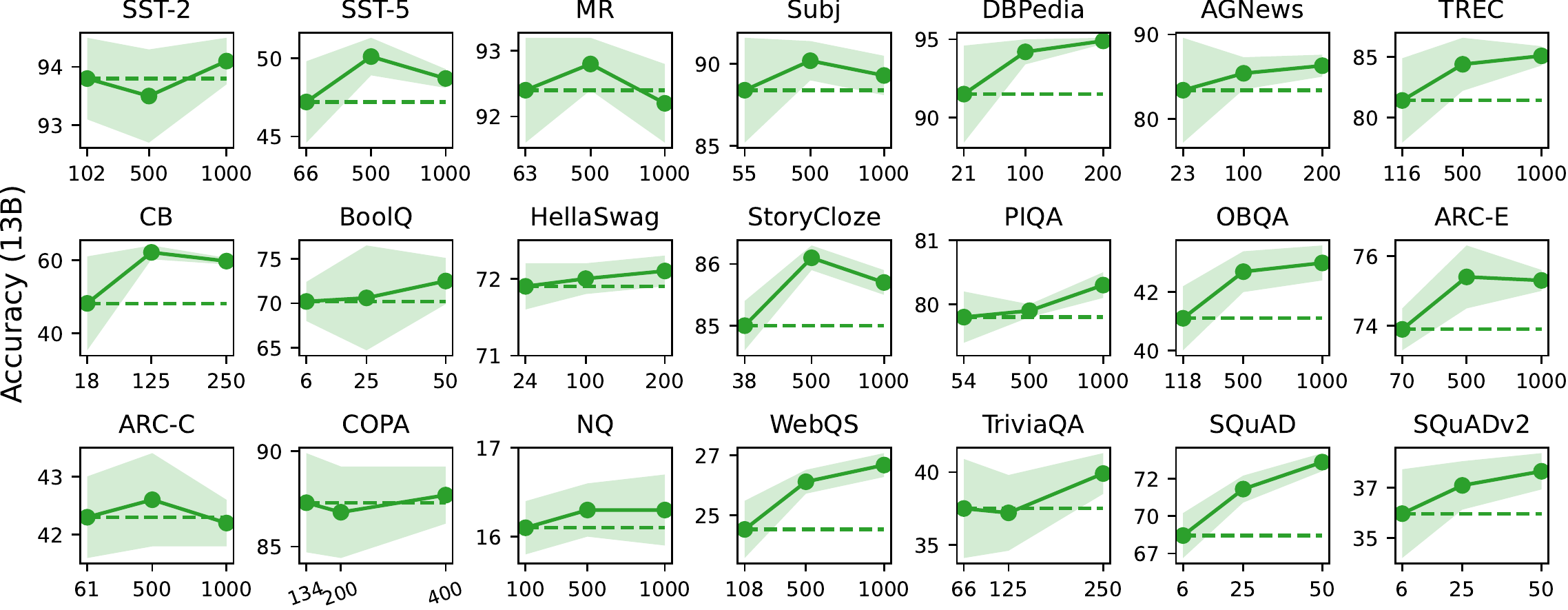}
\end{subfigure}
\caption{In-context learning performance with respect to the number of conditioning examples. The dotted line indicates the accuracy of the maximum shot that fills the entire context window. After adopting structured prompting, we can scale in-context learning to thousands of examples with better performance and lower variance.}
\label{fig:result}
\end{figure}

\subsection{Results}
\paragraph{Text Classification}
First, we consider text classification tasks, the results of nine datasets are shown in Table~\ref{tbl:classification}.
Conditioning on thousands of examples, structured prompting brings consistent and significant improvements (3-5 absolute gains) for in-context learning.
Moreover, our method makes in-context learning much more stable across multiple seeds, while conventional in-context learning is sensitive to different demonstration selections and permutations.
In most cases, providing more examples leads to better performance and lower variance.
For easier tasks like sentiment (SST-2, Subj, and MR) and topic classification (TREC and DBPedia), the improvement is more obvious and stable (the variance is generally less than 1.0).
For natural language inference (CB), the result is relatively unstable, e.g., the 6.7B model has an outlier that classifies all examples into the same class, which indicates that inference is still a challenging task for in-context learning.

\paragraph{Multi-Choice Tasks}
The performance comparison of multi-choice tasks is shown in Table~\ref{tbl:multichoice}.
Structured prompting still brings consistent gains on these tasks.
However, we notice that the improvement of our method on these tasks is relatively small compared with text classification. Besides, utilizing more demonstrations does not always lead to better performance.
Scaling up the model size instead of the number of demonstrations can bring more improvements in these tasks.

\paragraph{Open-Ended Generation}
Both text classification and multi-choice tasks restrict the label space.
To evaluate structured prompting on open-ended generations tasks, we consider two types of datasets: closed-book question answering without conditioning on auxiliary information and extractive reading comprehension.
Results are shown in Table~\ref{tbl:generation}.
We observe that for all datasets except Natural Questions, incorporating more demonstrations via structured prompting leads to a monotonously increasing performance boost and a decreasing variance.
Especially for SQuAD, our method has nearly five points improvement over the baseline on 13B LM with only 50 examples.
For NQ, there is a negative observation that structured prompting has little gains compared with conventional in-context learning.
For SQuAD, we tried up to 50-shot settings because of the long length of a single demonstration.
We believe that the performance can still increase if the number of demonstrations can be further expanded. 

\subsection{Scale up to 176-Billion Model}

The previous results show that the gains from structured prompting decrease slightly as the model size increases.
To verify the effectiveness of our method on huge models, we conduct experiments with a subset of datasets on BLOOM-176B~\citep{bloom}.\\
\paragraph{Evaluation Protocol}
Our experiments are implemented on 8$\times$80GB A100 GPUs. 
We evaluate conventional prompting under different prompt lengths (0.5$\times$ means the prompt fills half of the context window size) and structured prompting under different group numbers (5$\times$ means five groups of prompts).
Under each setting, we use five different random seeds for all tasks and report the average results.

\paragraph{Large Model Results}
The performance comparisons on 176-Billion Model are shown in Table~\ref{tbl:bloom}. 
For most datasets, the performance gets better as the number of groups increases. 
The variance results also show that structured prompting is highly stable when using five groups. 
For TREC and PIQA, we observe that 3$\times$ achieves better results that 5$\times$ but they both outperform the conventional in-context learning.
Our experiments show that large language models still have the potential to achieve better prompting results when utilizing more demonstrations.

\begin{table}[ht]
\resizebox{\textwidth}{!}{%
\begin{tabular}{lccccccccccl}
\toprule
\multicolumn{12}{c}{\emph{BLOOM-176B}} \\ \midrule
      & RTE        & CB          & SST-2      & SST-5      & TREC       & BoolQ      & Subj       & MR         & OBQA       & \multicolumn{1}{c|}{PIQA}       & Avg  \\ \midrule
    0$\times$    & 58.1$_{0.0}$          & 39.3$_{0.0}$          & 83.1$_{0.0}$          & 35.0$_{0.0}$          & 14.4$_{0.0}$          & 62.2$_{0.0}$          & 54.7$_{0.0}$          & 63.5$_{0.0}$          & 41.0$_{0.0}$          & \multicolumn{1}{c|}{78.5$_{0.0}$}          & 53.0          \\
0.25$\times$ & 67.8$_{4.9}$          & 53.9$_{27.5}$         & 94.1$_{1.1}$          & 43.9$_{6.2}$          & 51.8$_{4.8}$          & 72.7$_{2.7}$          & 88.5$_{6.3}$          & 92.7$_{0.4}$          & 44.0$_{0.8}$          & \multicolumn{1}{c|}{79.2$_{0.3}$}          & 68.9          \\
0.5$\times$  & 66.5$_{5.4}$          & 62.5$_{10.0}$         & 94.7$_{0.4}$          & 41.1$_{3.7}$          & 61.2$_{3.3}$          & 74.7$_{1.5}$          & 90.7$_{4.3}$          & 92.6$_{0.5}$          & 44.7$_{1.3}$          & \multicolumn{1}{c|}{79.8$_{0.3}$}          & 70.9          \\
1$\times$    & 67.1$_{7.7}$          & 75.4$_{5.6}$          & 94.2$_{0.5}$          & 46.7$_{2.0}$          & 88.6$_{2.3}$          & 75.5$_{2.2}$          & 93.6$_{1.2}$          & 92.6$_{0.4}$          & 45.4$_{1.1}$          & \multicolumn{1}{c|}{79.8$_{0.4}$}          & 75.9          \\
3$\times$    & 70.5$_{2.7}$          & 76.4$_{3.2}$          & 94.4$_{0.5}$          & 45.7$_{4.4}$          & \textbf{91.0$_{2.2}$} & 77.1$_{1.8}$          & 95.2$_{0.5}$          & 93.0$_{0.4}$          & 45.6$_{0.6}$          & \multicolumn{1}{c|}{\textbf{80.3$_{0.4}$}} & 76.9          \\
5$\times$    & \textbf{72.3$_{2.6}$} & \textbf{77.5$_{2.4}$} & \textbf{94.7$_{0.4}$} & \textbf{47.0$_{2.2}$} & 89.7$_{2.4}$          & \textbf{77.9$_{1.7}$} & \textbf{95.7$_{0.4}$} & \textbf{93.1$_{0.3}$} & \textbf{46.0$_{0.3}$} & \multicolumn{1}{c|}{80.1$_{0.2}$}          & \textbf{77.4} \\ \bottomrule

\end{tabular}%
}

\vspace{0.5cm}

\resizebox{\textwidth}{!}{%
\begin{tabular}{lcccccccccc}
\toprule                                                                            
      & RTE           & CB            & SST-2         & SST-5         & TREC          & Subj          & MR            & BoolQ         & OBQA          & PIQA        \\ \midrule
\#shot $(\!1\!\times\!)$    & 24          & 18          & 102          & 66          & 116          & 55          & 63          & 6          & 118          & 54         \\  \bottomrule
\end{tabular}%
}
\vspace{0.2cm}
\caption{Results averaged over 5 random seeds on BLOOM-176B. The table below shows the actual shot number that fills the entire context window. The shot number of other settings can be calculated by N$\times$\#shot. }
\label{tbl:bloom}

\end{table}

\subsection{Stability Analysis}
Prior efforts demonstrate that different choices and permutations of conditioning examples can cause a high variance for in-context learning.
We now investigate their impacts on structured prompting.
Figure~\ref{fig:result} shows how the performance and variance of in-context learning change as the number of examples increases.
We observe that our method can significantly reduce inference variance for text classification and open-ended generation tasks.
This phenomenon is less pronounced for multi-choice tasks that correlate better with pretraining.
For larger LMs (13B), the baseline approach is stable enough in the max-shot case.
Structured prompting can further boost its stability while achieving better performance.
It suggests that in-context learning is underestimated under the few-shot setting and our method can bring key benefits to maximize its stability and effectiveness.

\subsection{Ablation Studies}
\begin{figure}[ht]
 \centering
 \begin{subfigure}[b]{0.48\textwidth}
     \centering
     \includegraphics[width=\textwidth]{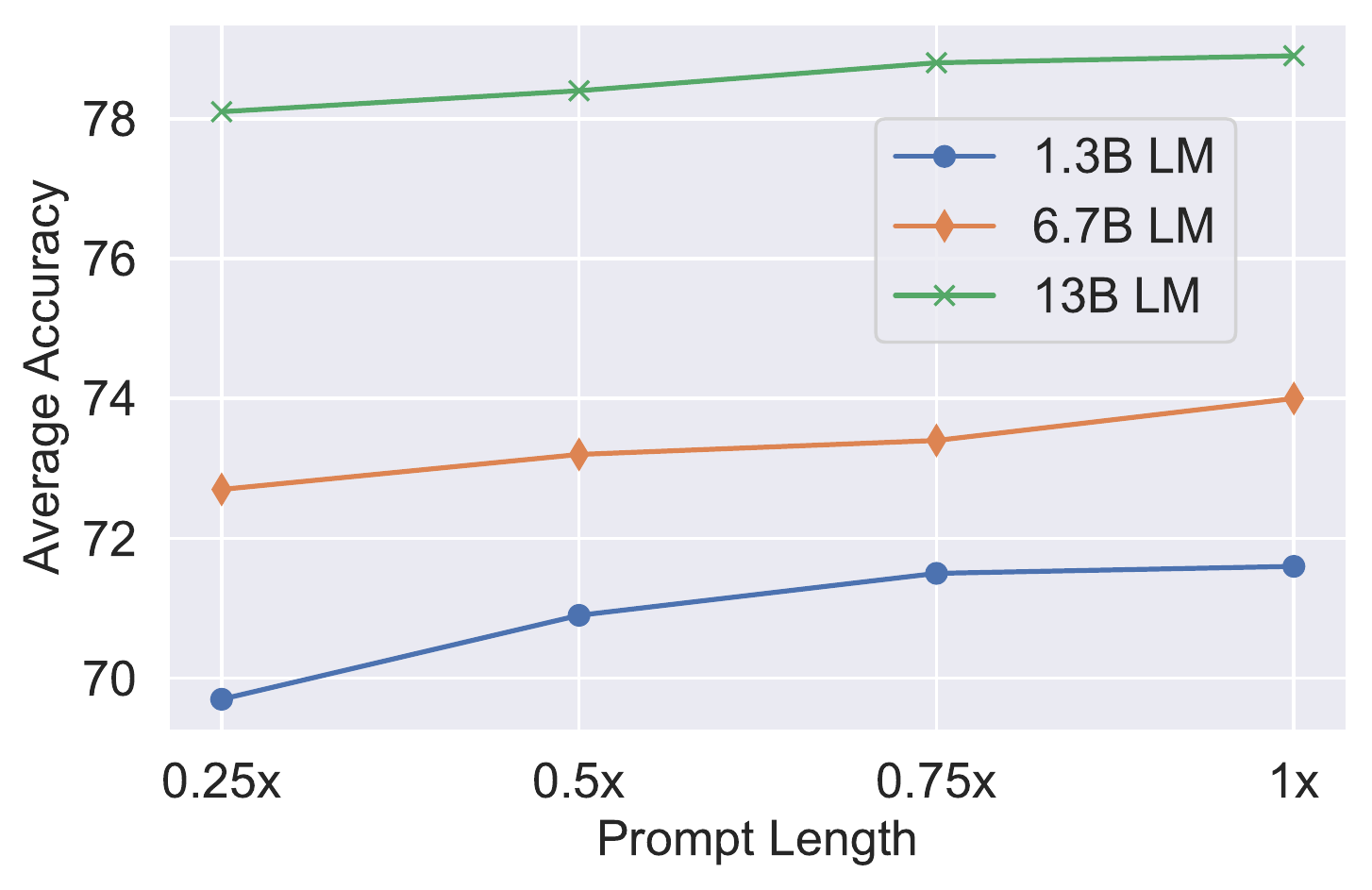}
      \caption{Ablation of Prompt Length}
     \label{fig:len_ablation}
 \end{subfigure}
 \hfill
 \begin{subfigure}[b]{0.48\textwidth}
     \centering
     \includegraphics[width=\textwidth]{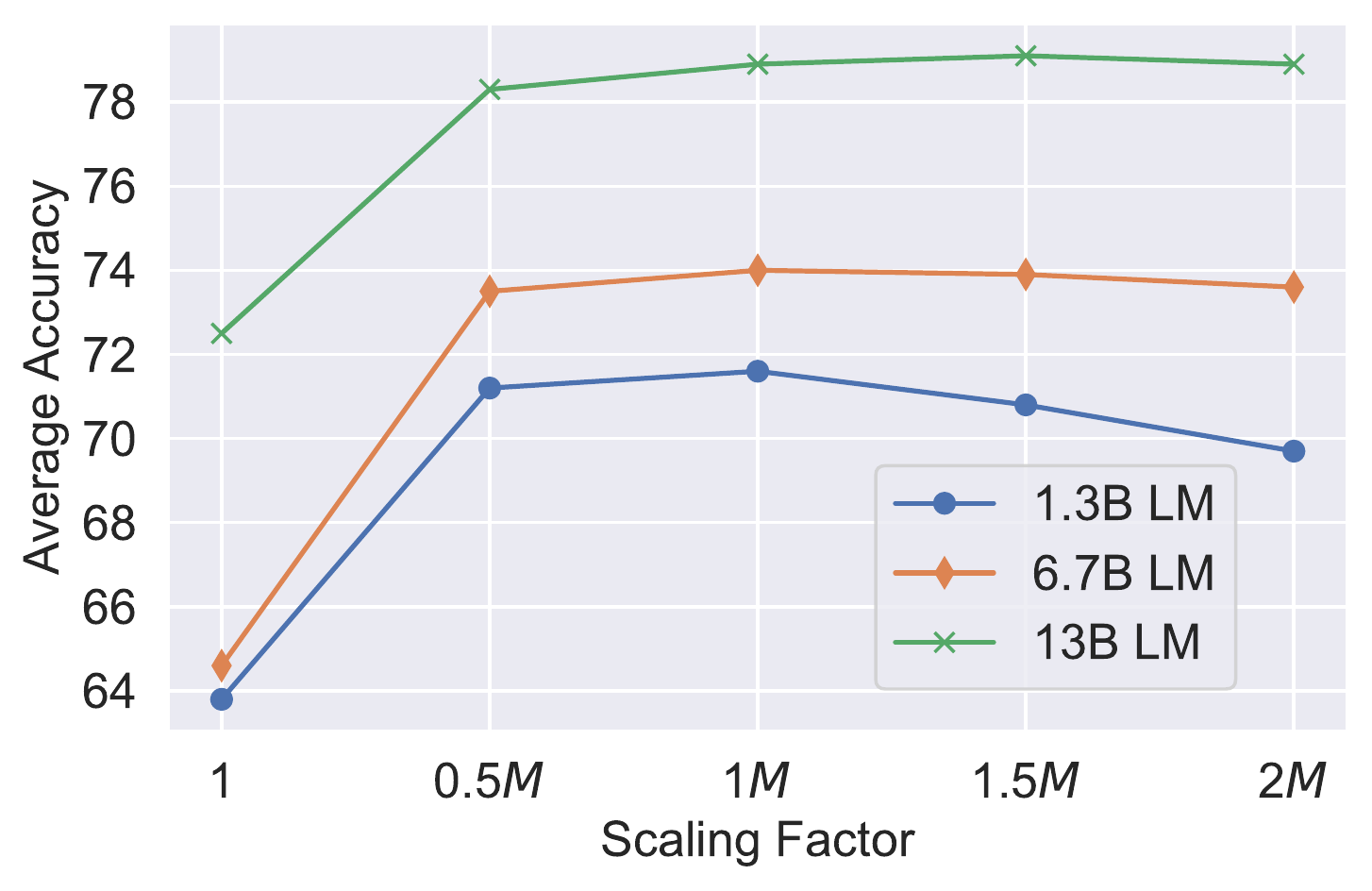}
      \caption{Ablation of Scaling Factor}
     \label{fig:scalar_ablation}
 \end{subfigure}
\caption{Ablations of structured prompting.}
\end{figure}

\paragraph{The Effect of Prompt Length}
The results of different prompt lengths are shown in Figure~\ref{fig:len_ablation}. We control the number of examples as a constant value, so the group number is inversely proportional to the prompt length. Generally, the longer prompt length means better accuracy. In the small model (1.3B), the performance has a big drop when the prompt length is small (0.25$\times$). It shows that the auto-regressive structure is still the most ``natural'' one. The best way to use structured prompting is to expand the groups under the maximal sequence length in pre-training. Therefore, we believe that the fundamental problem is language models' ability to deal with exceeding sequence length. If the model's extrapolation performance is satisfying, the benefits of structured prompting will degenerate to memory saving. We leave it for future work. 

\paragraph{The Effect of Scaling Factor}
The scaling factor is essential in structured prompting. The results of various scaling factors are shown in Figure~\ref{fig:scalar_ablation}. We observe that without the scaling factor, the attention distribution will focus on the demonstration, leaving the query alone. As illustrated before, a natural way is to repeat the query the same times as the group number. With that, the query tokens seem to be concatenated with every demonstration. The experiments show that the appropriate scaling factor contributes to huge progress compared with the naive inference. Besides, the exact M multiplier is best among the surrounding values, although large language models are more robust to the disturbance of the scaling factor.

\paragraph{The Effect of Alignment Strategies}
In structured prompting, the group should have the same length to ensure the query and demonstrations are continuous. In one group, we initialize tokens by padding the length which is set as a constant. Then, we fill with examples sequentially from right to left. When the vacancy's length is smaller than the incoming example's length, three ways are optional to deal with the left padding:
\begin{itemize}
    \item \textbf{Attention Mask}: Masking the padding tokens during the whole process, including the calculation for ($K_{\mathcal{Z}_i}$, $V_{\mathcal{Z}_i}$) and the attention stage at inference step.
    \item \textbf{Pad Space}: Replacing the padding tokens with blank space tokens. In this case, the attention is calculated in every token.
    \item \textbf{Truncate}: filling the padding tokens with the incoming example but the front is truncated to maintain the constant length.
\end{itemize}

Table~\ref{tbl:strategy} shows that the ``Truncate'' strategy works well for both models. In FairseqLM, there is a \texttt{<bos>} token so that the subsequent tokens are less disturbed. In BLOOM, transitional invariance of Alibi~\citep{alibi} is used to deal with padding and mask. However, it has a significant drop with ``Pad Space''. The absence of \texttt{<bos>} seems to amplify the noise brought by blank space tokens. In conclusion, the ``Truncate'' strategy is the easiest and most natural way for aligning groups.

\begin{table}[ht]
\centering
\begin{tabular}{lcccccc}
\toprule
\textbf{Strategies} & \textbf{TREC}                  & \textbf{BoolQ}                 & \textbf{DBPedia}               & \textbf{AGNews}                & \multicolumn{1}{c|}{\textbf{HellaSwag}}             & \textbf{Average} \\
\midrule
\multicolumn{7}{c}{\emph{BLOOM-7B}} \\ \midrule
w/o Right-Alignment &  56.9 & \textbf{62.2} & 94.7 & 82.1 & \multicolumn{1}{c|}{\textbf{58.6}} & 70.9 \\
Attention Mask & \textbf{61.9}         & \textbf{62.2}          & 94.5          & 83.4          & \multicolumn{1}{c|}{\textbf{58.6}} & 72.1          \\
Pad Space      & 58.0          & \textbf{62.2}        & 93.1         & \textbf{84.8} & \multicolumn{1}{c|}{58.1}          & 71.2          \\
Truncate       & \textbf{61.9} & \textbf{62.2} & \textbf{95.4} & 84.3        & \multicolumn{1}{c|}{58.5}          & \textbf{72.5} \\
\midrule
\multicolumn{7}{c}{\emph{FairseqLM-6.7B}} \\ \midrule
w/o Right-Alignment &  79.0 & 61.4 & 92.8 & \textbf{75.4} & \multicolumn{1}{c|}{68.3} & 75.4 \\
Attention Mask &  79.3 & 61.9 & 92.9 & \textbf{75.4} & \multicolumn{1}{c|}{\textbf{68.7}} & 75.6 \\
Pad Space      & \textbf{80.0} & \textbf{62.0} & 93.0 & 74.7 & \multicolumn{1}{c|}{\textbf{68.7}} & \textbf{75.7} \\
Truncate       & 79.3 & 61.7 & \textbf{93.8} & 74.3 & \multicolumn{1}{c|}{\textbf{68.7}} & 75.6 \\
\bottomrule
\end{tabular}
\vspace{0.2cm}
\caption{Comparisons of different strategies for aligning multiple prompts.
FaiseqLM is insensitive to alignment strategy, while BLOOM prefers ``Attention Mask'' and ``Truncate''.}
\label{tbl:strategy}
\end{table}

\section{Related Work}
\label{related}
\paragraph{Improving In-Context Learning}
Despite surprisingly effective, in-context learning suffers from certain vulnerabilities.
For instance, the order of demonstrations and the choice of templates can cause a high variance in performance.
\citep{zhao2021calibrate} show that the variance arises because of three types of biases (majority label, recency, and common token bias) and propose to calibrate model prediction by content-free output.
\citep{protocalibration} demonstrate that these biases cause the decision boundary shift and propose calibrating it by estimating the distribution of prototypical clusters.
Other work focus on prompt engineering, including selecting the performant demonstration permutation~\citep{lu2022order} and semantically-similar in-context examples with a retrieval module~\citep{liu2021makes, learningtoretrieval}.
We aim to improve in-context learning by scaling up the number of demonstrations.

\paragraph{Understanding In-Context Learning}
Another line of work investigates understanding how in-context learning works. 
\citep{bayesian} propose a Bayesian inference framework to explain it, where the language model implicitly infers a concept when making a prediction.
Since in-context learning emerges after pretraining on a large corpus, some efforts study the correlation between pretraining corpus and in-context learning performance~\citep{corpora,freqimpact}.
Additionally, previous work~\citep{min2022rethinking,kim2022groundtruth} investigates whether the label-mapping of demonstrations matters as expected.

\paragraph{Fusion-In-Decoder}
\cite{fusionindecoder} propose Fusion-In-Decoder for encoder-decoder fine-tuning. The method was applied to open-domain question answering in order to leverage retrieved passages.
Specifically, each retrieved supporting passage is encoded by bidirectional encoders. Then the decoder performs conventional attention over the concatenation of the representations of passages.
In comparison, we focus on in-context learning with decoder-only models (such as GPT), without fine-tuning the original parameters.
There are also several key technical differences, which are critical to making the method work well.
First, we proposed rescaled attention to balance the attention allocation between context and test input.
Second, we right-align position embeddings for structured context so that they have the same relative distance with respect to the input.

\section{Discussion and Conclusion}

In this work, we explore how to utilize more examples for in-context learning and propose structured prompting to scale up the number of examples under restricted computation complexity.
We encode each group of demonstrations independently and prompt the language model with the concatenations of their representations via the rescaled attention.
Experimental results across a diverse set of tasks show that our method outperforms the conventional approach.
As the number of examples increases, our method achieves further gains and is much more stable.

Despite the promising results, the current method still has some limitations.
Ideal in-context learning should be invariant to demonstration permutations. 
If our method ensures that each group only contains one example, it satisfies the property indeed.
However, in our experiments, we find that it works well on smaller models (i.e., 1.3B) but does not work on larger models (i.e., 13B).
We hypothesize that larger models benefit more from autoregressive information, so we include multiple examples in each group.
For future work, we will dive more deeply into this direction.
Moreover, there is a mismatch between patterns of language model pretraining and in-context learning.
The current objective makes language models only aware of sequential relationships but not parallel relationships. 
We would like to incorporate this prior knowledge during pretraining so that it aligns better with the scheme of downstream inference.
In addition, structured prompting can be used to inject many long documents as context, e.g., using retrieved texts to augment generation.

\bibliographystyle{alpha}
\bibliography{sprompt}

\newpage
\appendix

\section{Templates}
\label{app:template}

\begin{table}[h!]\small
\centering
\begin{tabularx}{\textwidth}{lXX}

\toprule
\textbf{Dataset} & \textbf{Template} & \textbf{Label Space} \\
\midrule

\multicolumn{3}{c}{\emph{Text Classification}} \\ \midrule

SST-2 & Sentence: \{Sentence\} & Negative / Positive \\

&Label: \{Label\} & \\
\midrule

SST-5 & Sentence: \{Sentence\} & terrible / bad / neutral / good / great \\
&Label: \{Label\} & \\

\midrule

MR & Review: \{Sentence\} & Negative / Positive\\
&Sentiment: \{Label\} & \\

\midrule

Subj & Input: \{Sentence\} & objective / subjective \\
&Type: \{Label\} & \\

\midrule

DBPedia & Input: \{Sentence\} \newline Type: \{Label\}  & company / school / artist / athlete / politics / transportation / building / nature / village / animal / plant / album / film / book \\
\specialrule{0em}{2pt}{2pt}

\midrule

AGNews & Classify the news articles into the categories of World, Sports, Business, and Technology. & World / Sports / Business / Technology \\
\specialrule{0em}{2pt}{2pt}
&News: \{Sentence\} & \\
&Type: \{Label\} & \\

\midrule

TREC & Question: \{Sentence\} \newline Type: \{Label\}  & Description / Entity / Expression / Person / Number / Location \\

\midrule

CB & \{Premise\} & True / False / Neither \\
&Question: \{Hypothesis\} True, False, or Neither? & \\
&Answer: \{Label\} & \\

\midrule
BoolQ & Passage: \{Passage\} & No / Yes \\
&Question: \{Question\} Answer: & \\

\midrule

RTE & Passage: \{Premise\} & Yes / No \\
&Question: \{Hypothesis\} Yes or No? & \\

\midrule

\multicolumn{3}{c}{\emph{Open-Ended Generation}} \\ \midrule

NQ & Question: \{Question\} Answer: & Open-ended \\ \midrule

WebQS & Question: \{Question\} Answer: & Open-ended \\ \midrule

TriviaQA & Question: \{Question\} Answer: & Open-ended \\ \midrule

SQuAD & Passage: \{Passage\} & Open-ended \\ 
&Question: \{Question\} Answer: & \\ \midrule

SQuADv2 & Passage: \{Passage\} & Open-ended (The golden label is ``none'' if the \\ 
&Question: \{Question\} Answer: & question is answerable) \\ 

\bottomrule
\end{tabularx}
\vspace{0.2cm}
\caption{Prompt template and label mapping for text classification and open-ended generation tasks in our experiments. For multi-choice tasks, we use the same protocol as GPT-3, i.e., choose the best completion of a given prefix.}
\label{apptable1}
\end{table}

\end{document}